\PassOptionsToPackage{numbers}{natbib}
\PassOptionsToPackage{dvipsnames}{xcolor}
\documentclass{article}

\usepackage[final]{neurips_2022}

\usepackage[utf8]{inputenc} 
\usepackage[T1]{fontenc}    
\usepackage[hidelinks]{hyperref}     
\usepackage{url}            
\usepackage{booktabs}       
\usepackage{amsfonts}       
\usepackage{nicefrac}       
\usepackage{microtype}      
\usepackage[dvipsnames]{xcolor}         
\usepackage[pdftex]{graphicx}
\usepackage{subfig}
\usepackage{enumitem}
\usepackage{threeparttable, booktabs, multirow, diagbox}
\usepackage{siunitx}
\usepackage{mathrsfs}
\usepackage{wrapfig}

\usepackage{prettyref}
\newrefformat{sec}{Section~\ref{#1}}
\newrefformat{fig}{Fig.~\ref{#1}}
\newrefformat{table}{Table~\ref{#1}}
\newrefformat{appendix}{Appendix~\ref{#1}}

\newcommand\boldred[1]{\textcolor{BrickRed}{\textbf{#1}}}

\newcommand\boldgreen[1]{\textcolor{ForestGreen}{\textbf{#1}}}

\definecolor{darkpastelgreen}{rgb}{0.01, 0.75, 0.24}

\let\subparagraph\paragraph
\usepackage[compact]{titlesec}

\titlespacing{\section}{0pt}{1ex}{1ex}
\titlespacing{\subsection}{0pt}{1ex}{0.5ex}
\titlespacing{\subsubsection}{0pt}{0.5ex}{0.5ex}

\DeclareMathSizes{9}{8.2}{5}{3}
\title{Towards Continually Learning Application Performance Models}

\author{%
  Ray A. O. Sinurat\\
  Department of Computer Science\\
  University of Chicago\\
  \texttt{rayandrew@cs.uchicago.edu} \\
  \And
  Anurag Daram  \\
  Department of Electrical \& Computer Engineering  \\
  University of Texas at San Antonio \\
  \texttt{anurag.daram@utsa.edu} \\
  \AND
  Haryadi S. Gunawi \\
  Department of Computer Science\\
  University of Chicago\\
  \texttt{haryadi@cs.uchicago.edu} \\
  \And
  Robert B. Ross\\
  Mathematics \& Computer Science Division\\
  Argonne National Laboratory \\
  \texttt{rross@mcs.anl.gov} \\
  \And
    Sandeep Madireddy\thanks{Contact for Correspondence}   \\
  Mathematics \& Computer Science Division\\
  Argonne National Laboratory \\
  \texttt{smadireddy@anl.gov} \\
}

\begin{document}

\maketitle

\begin{abstract}

    Machine learning-based performance models are increasingly being used to build critical job scheduling and application optimization decisions. Traditionally, these models assume that data distribution does not change as more samples are collected over time. However, owing to the complexity and heterogeneity of production HPC systems, they are susceptible to hardware degradation, replacement, and/or software patches, which can lead to drift in the data distribution that can adversely affect the performance models. 
    To this end, we develop continually learning performance models that account for the distribution drift, alleviate catastrophic forgetting, and improve generalizability.
    Our best model was able to retain accuracy, regardless of having to learn the new distribution of data inflicted by system changes, while demonstrating a 2$\times$ improvement in the prediction accuracy of the whole data sequence in comparison to the naive approach.  
    
\end{abstract}

\section{Introduction}

The complexity of leadership-class high-performance computing (HPC) systems is increasing rapidly due to the need to handle diverse workloads and applications. In particular, storage systems and I/O architectures are integrating different heterogeneous storage technologies to maximize the price-performance trade-off. This complexity entails the need for sophisticated empirical/machine learning models to accurately predict the application performance. The accuracy of the performance model is crucial due to its use in making decisions about job scheduling, application optimization, and capacity planning of facilities.

It is commonly assumed that the data used to learn the performance models do not undergo any distribution shift. Under this stationarity assumption, the observation of more data on the application performance in the HPC system should increase the predictive performance of these machine learning models. However, as reported in recent works~\cite{mundt2020wholistic}, factors such as hardware degradation~\cite{Kasick2010-vv}, replacement, anomalies~\cite{Gunawi2018-xc} or software upgrades~\cite{Lockwood2018-cl} can affect the state of the system and, in turn, lead to a change/drift in the underlying distribution. Performance models trained on the past data will be degraded due to this data distribution shift.  
Such distribution shifts have been handled either by updating the model ad hoc without drift detection~\cite{Gama2014-nd} using a sliding window of data or by ignoring those data before the drift occurred. More recently, \citet{Madireddy2019-tv} proposed a moment-matching transform to correct for data drift post-detection. 
A more general approach to train/adapt the performance model in the presence of this drift is using \emph{continual learning}~\cite{Thrun2012-px} algorithms, which acknowledge the presence of data distribution shifts and are designed to prevent catastrophic forgetting of the models on the data observed before the shift when training on data observed post-drift, thus generalizing across the distribution shifts. The ability to learn continuously from an incoming data stream without catastrophic forgetting is critical for designing intelligent systems. 


However, prior research in continuous learning has focused mainly on 
\emph{virtual concept drift}~\cite{Lesort2021-sc} (or label drift) where there exists drift in the distribution of targets (\(P(y)\)) without affecting the functional relationship between
the inputs and outputs (\(P(y|x)\)) of the model. For example, this scenario is encountered in training classification models, which should learn a new class without forgetting the previous ones when presented sequentially, one after the other.
However, for performance modeling, we are interested in the \emph{real concept drift} scenario in which learning occurs in a sequence of tasks where the input distribution (\(P(x)\)) remains the same but the functional relationship (\(P(y|x)\)) changes across tasks due to the change in the state of the system. This scenario has been less explored with few studies pertaining to image segmentation~\cite{Cermelli2020-ls} and improvement in label precision~\cite{Abdelsalam2021-ne}. 

To this end, we make the following contributions to this work: \textbf{(i)} we formulate performance modeling in the presence of data drifts as a \emph{real concept drift} continuous learning scenario; \textbf{(ii)} adapted several \emph{virtual concept drift} continuous learning approaches to performance modeling and compared to naive learning that ignores catastrophic forgetting; finally, \textbf{(iii)} our results show a \textbf{\(\textbf{2}\)-fold} improvement in the accuracy of the continual learning models compared to their naive learning counterparts after learning all tasks.




\section{I/O Performance Modeling}
\label{sec:io-perf-modelling}

\paragraph{Data Preparation:}

\renewcommand{\arraystretch}{1.5}
 \begin{wraptable}{r}{70mm}
\vspace{-0.4in}
\caption{
Input features extracted from the I/O monitoring data.}
\centering
\label{tab:inp_feat}{}
   \begin{threeparttable}[t]
     \resizebox{0.5\textwidth}{!}
     {
\begin{tabular}{clcc}
\hline
\textbf{Metric} & \multicolumn{1}{c}{\textbf{Description}} & \textbf{Units} & \textbf{Tool} \\ \hline
\hline

Perc\_OST\_Full & Average $\%$ OST fullness  & -- & LMT\\ \hline
Ave\_OSS\_CPU  & Average OSS CPU load  & --  & LMT \\ \hline
Ave\_MDS\_CPU  & Average MDS CPU Load  & --  & LMT \\ \hline
Num\_Conc\_Jobs & Number of concurrent jobs  & -- & Slurm \\ \hline
FS\_Read\_Vol & Total read volume across FS  & GB & LMT \\ \hline
FS\_Read\_Vol & Total write volume across FS & GB & LMT \\ \hline
Num\_Mkdir\_Op & Number of mkdir operations  &  --  & LMT \\ \hline
Num\_Rename\_Op & Number of rename operations&  --  & LMT \\ \hline
Num\_Rmdir\_Op & Number of rmdir operations  &  --  & LMT \\ \hline
Num\_Unlink\_Op & Number of unlink operations &  --  & LMT \\ \hline
    			\bottomrule
\end{tabular} 
}
\end{threeparttable}

\end{wraptable}

We collected our data from Cori, a production supercomputing system at the National Energy Research Scientific Computing Center that is used to run high-performance applications (HPC) tasks. Cori is a Cray XC40 system that comprises of 12,000 compute node and luster file systems with a 30PB capacity and peak performance of 700 GB/sec. The data used in our experiments consist of eight production applications with diverse application workloads representative of the science application typically run.  We used TOKIO (Total Knowledge of I/O; \cite{Lockwood2018-aw}) framework for I/O performance profiling. As part of TOKIO, the application performance statistics is provided by Darshan~\cite{Carns2011-vc} logs, I/O traffic in Lustre file systems is 
obtained using LMT~\cite{Garlick_undated-yd} (Lustre Monitoring Tools), and scheduling information using Slurm~\cite{Yoo2003-dx}.

\setlength\intextsep{2pt}
\begin{wrapfigure}{r}{70mm}
\centering
  \includegraphics[width=\linewidth]{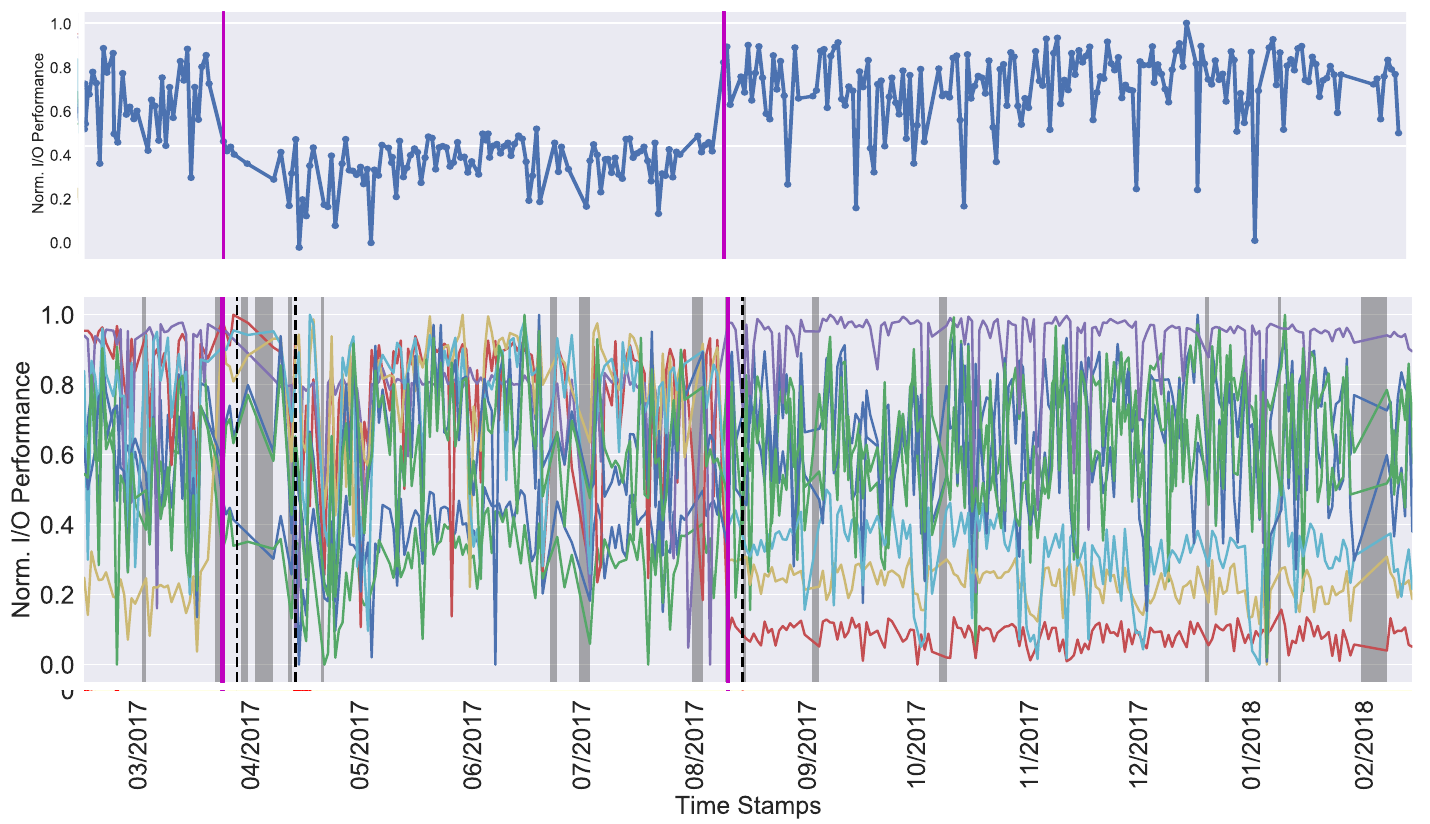}
\caption{I/O performance as a function of time for all eight applications (bottom) and one of the applications (top); Magenta lines show the location of the software upgrade.}
\label{fig:data}
\vspace{-0.20in}
\end{wrapfigure}

All of these applications were run contemporaneously on Cori; hence, the temporal location of the system changes should be consistent across the applications. Specifically, two major software upgrades were performed during the course of this study. Both upgrades affected the I/O performance of some applications.
Therefore, it can be assumed that the data are divided into three different regimes by
the two software upgrades as seen in Table~\ref{fig:data} that show the normalized I/O performance for a single application at the top and all eight applications at the bottom. 
The magenta line indicates the time that the software updates were performed, and the top figure shows the shift in the I/O performance distribution. We consider metrics that capture filesystem traffic (using LMT) and system load (using Slurm) as features used by the performance model to predict I/O performance. We also do not preprocess these features to choose a subset of these metrics that are not collinear. A summary of the adopted features is shown in Table~\ref{tab:inp_feat}.



\section{Continual Learning}

The goal of continual learning (CL) is to improve the model over time while retaining prior knowledge or experiences. CL came to fruition as a solution when the model forgets prior knowledge due to a dynamic data stream that changes over time \cite{Lesort2021-sc}. This change is typically in the data distribution which leads to data drift/non-stationarity and ignoring it can lead to catastrophic forgetting \cite{French1999-sv}. Formally, continual learning on a sequence of tasks (\(T^t, t=1,2,..,N\) $\forall \, T \in \mathscr{T}$) consisting of ordered pairs of input data points and their corresponding targets \{\(\mathcal{X}^t,\mathcal{Y}^{t}\)\} aims to maximize the performance of a system across all tasks \(\mathscr{T}\), when trained sequentially. There are different kinds of data drifts that introduce non-stationarity in learning and motivate the need for continual learning approaches. Some well-known drifts that affect supervised learning include:
(1) \textbf{real concept drift}: occurs when the distribution of inputs remain same across tasks, i.e, \(P_t(x) = P_{t+1}(x), \forall x \in \mathcal{X} \) but the functional relation changes, i.e., \(P_t(y|x) \neq P_{t+1} (y|x), \forall (x,y) \in (\mathcal{X},\mathcal{Y})\); (2) \textbf{virtual concept drift} happens when the distribution shifts only happen inside target variables (\(P_t(y) \neq P_{t+1} (y)\)) without altering the relationship towards input variables; (3) \textbf{domain drift} exists if the drift occurs in the input variables (\(P_t(x) \neq P_{t+1} (x)\)) without affecting the relationship towards target variables. 

We observed that the I/O performance data undergoes \emph{real concept drift}, which is much harder and less studied in the context of continual learning. Most works have considered \emph{virtual concept drift} in the classification scenario where data from disjoint classes form new tasks. Although I/O performance modeling is a regression problem, we formulate it as a classification problem by dividing the continuous output range into an equally spaced interval for simplicity, in addition to the potential to adapt the high-performing algorithms from the literature as well as software that have been primarily targeting the classification scenarios.




\textbf{Formulation and implementation of learning with real concept drift:}
To this end, we use the API provided by Avalanche \cite{avalanche}, a continual learning framework based on PyTorch \cite{Paszke2019-cr}, and adopt the existing implementation of the classification-based continual learning algorithms to learn in the presence of real concept drift. Specifically, we incorporated the class-incremental setting \cite{Masana2020-ln} from \emph{virtual concept drift} works but modified the learning so that each task's output spans the entire space as opposed to each of them accumulating from disjoint classes in the previous case. In addition to that, we keep the task id in training and inference consistent with the traditional class-incremental learning, where the task id is provided at training, but not inference. In this paper, the task id is an integer that provides essential information when the software upgrade occurs.
We also note that as we described in \prettyref{sec:io-perf-modelling}, we used the aggregation of performance statistics provided by Darshan \cite{Carns2011-vc} as our target variable (\(y\)) which results in a single floating number that should be restricted to regression problems. We converted regression problems into classification problems by splitting the \(y\), which infinitely spans from \([0.0,1.0]\), into 10 regimes equally.


To train our model, we need to provide features (\(x\)), targets (\(y\)) at training time, and additional task id (\(t\)) while splitting the data. We achieve this by providing task id (\(t\)) to Avalanche data loader and treating each dataset with a different task id as a separate dataset. This separation of the dataset is required for informing Avalanche to split the data correctly based on its performance shifts after upgrading software.
Our model is implemented in PyTorch which then will be continually learning inside Avalanche based on the chosen methodology. 



\textbf{Continual learning strategies:} In this work, we adopt six different methodologies that are popular benchmarks and have an existing implementation within Avalanche \cite{avalanche}. These approaches include (1)
\textbf{Elastic Weight Consolidation (EWC)} \cite{kirkpatrick_2017_Overcoming}, which is a regularization-based approach that measures the importance of the parameters for the current task and penalizes future updates. The regularization term of EWC  consists of a quadratic penalty term for each previously learned task, whereby each task’s term penalizes the parameters for how different they are compared to their value directly after finishing the training on that task. The strength of each parameter’s penalty depends for every task on how important that parameter was estimated to be for that task, with higher penalties for more important parameters; (2) \textbf{Synaptic Intelligence (SI)}~\cite{zenke2017continual} is another regularization-based approach that consists of only one quadratic term that penalizes changes to important parameters which are identified by tracking each synapse's credit assignment during the task. The importance parameter is measured by computing the per parameter contribution to the change of loss for the current task and thus strongly contributing parameters are heavily penalized in subsequent tasks. (3) \textbf{Learning without Forgetting (LwF)} \cite{li2017learning} is a distillation-based approach towards continual learning, wherein previous model outputs are used as soft labels for previous tasks. For this, each input to be replayed is labeled with a “soft target”, which is a vector containing a probability for each active class; (4) \textbf{Averaged Gradient Episodic Memory (A-GEM)} \cite{chaudhry2018efficient} uses episodic memory as an optimization constraint to avoid catastrophic forgetting. The sample handling of A-GEM avoids solving a quadratic optimization problem for handling samples in the buffer and, instead uses the mean gradient of such samples from the buffer. The sample is selected, if they point in the same direction in which the current gradient is applied, otherwise an orthogonal projection to the averaged gradient is performed; (5) \textbf{Gradient-based Sample Selection (GSS) Greedy}  \cite{aljundi2019gradient} is a sample selection strategy for a setup without
task boundaries or at least knowledge about these. Each seen
sample is regarded as an individual constraint, to which every
following sample must be compatible. Sample selection in this context is identical to a constraint reduction problem which is solved by a greedy strategy. This strategy selects \textit{n} random samples from the
buffer and calculates the cosine-similarity between the gradient
of the current sample and the gradients of the selected samples.
Samples of the buffer are only replaced if the similarity falls
below a defined threshold, that is, the sample with
maximal cosine-similarity is replaced; (6) \textbf{Greedy Sampler and Dumb learner (GDumb)} \cite{prabhu2020gdumb} greedily stores samples balanced over the observed classes and at test time learns a new model from scratch with the rehearsal memory. This approach consists of two components, namely the gradient balancing sampler and learner.  The sampler greedily creates a new bucket for that class and starts removing samples from the old ones, in particular, from the one with a maximum number of samples. 

We also adopt the baseline/\textbf{Naive}, which is a simple methodology in continual learning that does not do any optimization to prevent catastrophic forgetting. In Naive, a model is just incrementally fine-tuned depending on the new (drifted) data. Naive is provided natively inside Avalanche to give the lowest baseline of a model that suffers from catastrophic forgetting and mimics typical (stationary) supervised learning.

\section{Results and Discussions}

\renewcommand{\arraystretch}{1.0}
\setlength{\intextsep}{0pt}%
\begin{wraptable}{R}{0.465\linewidth}
        \vspace{-0.15in}
    	\caption{Accuracy and forgetting metrics.}
    	\centering
    	\label{table:results}
    	\begin{threeparttable}[t]
        \resizebox{0.45\textwidth}{!}
         {
    		\begin{tabular}{lcccc}
    			\toprule
    			Method & Task \# & Train Acc & Avg Acc & Avg Forgetting \\
    			\midrule
                \midrule
                
    			\multirow{3}{*}{Naive} & 1 & 0.668 & 0.589 & 0 \\ 
    			& 2 & 0.857 & 0.521 & 0.315 \\ 
    			& 3 & 0.879 & 0.367 & \boldred{0.531} \\
    			\hline
    			
    			\multirow{3}{*}{Synaptic Intelligence} & 1 & 0.668 & 0.589 & 0 \\ 
    			& 2 & 0.702 & 0.511 & 0.260 \\ 
    			& 3 & 0.776 & 0.338 & 0.503 \\
    			\hline
    		
    			\multirow{3}{*}{EWC} & 1 & 0.668 & 0.589 & 0 \\ 
    			& 2 & 0.850 & 0.546 & 0.260 \\ 
    			& 3 & 0.952 & 0.370 & 0.526 \\
    			\hline
    			
    			\multirow{3}{*}{LwF} & 1 & 0.668 & 0.589 & 0 \\ 
    			& 2 & 0.857 & 0.521 & 0.315 \\ 
    			& 3 & 0.879 & 0.367 & \boldred{0.531} \\
    			\hline
    			
    			\multirow{3}{*}{AGEM} & 1 & 0.668 & 0.589 & 0 \\ 
    			& 2 & 0.837 & 0.515 & 0.315 \\ 
    			& 3 & 0.924 & 0.395 & 0.489 \\
    			\hline

                \multirow{3}{*}{GSS Greedy} & 1 & 0.617 & 0.521 & 0 \\ 
    			& 2 & 0.984 & 0.734 & -0.192 \\ 
    			& 3 & 0.990 & \boldgreen{0.619} & 0.197 \\
    			\hline
    			
    			\multirow{3}{*}{GDumb} & 1 & 0.694 & 0.589 & 0 \\ 
    			 & 2 & 0.712 & 0.596 & 0.014 \\ 
    			 & 3 & 0.703 & \boldgreen{0.623} & -0.027 \\
    			
    			\midrule
    			\bottomrule
    		\end{tabular}
    	}
    	\end{threeparttable}
\end{wraptable}
The following section defines the experiment setup and establishes the improvement induced by the use of performance modeling as proposed. All experiments were carried out inside \emph{bare metal} machines with a single Intel (R) Xeon (R) CPU @ 2.00GHZ CPUs consisting of 8 cores with hyper-threading enabled, 12 GiB of memory, and 200 GB of disk space. The configurations of the model and the continual learning methodologies can be seen in \prettyref{appendix:hyperparams}. For memory-based methodologies such as GSS Greedy and GDumb, we fixed the memory size to obtain fair results and comparisons.
To evaluate the efficacy of our proposed model, we used two different metrics: (1) \textbf{Average Forgetting} that measures how bad a model forgets about prior tasks after being trained on a new task \cite{Mai2021-er}; and (2) \textbf{Average Accuracy} is used to determine the accuracy of the trained model to predict current and prior data after training \cite{Mai2021-er}.
\begin{wrapfigure}{R}{0.90\linewidth}
   \vspace{-0.1in}
   \centering
   \makebox[\linewidth][c]{
        \subfloat[\centering Average Forgetting]{\includegraphics[width=0.51\linewidth]{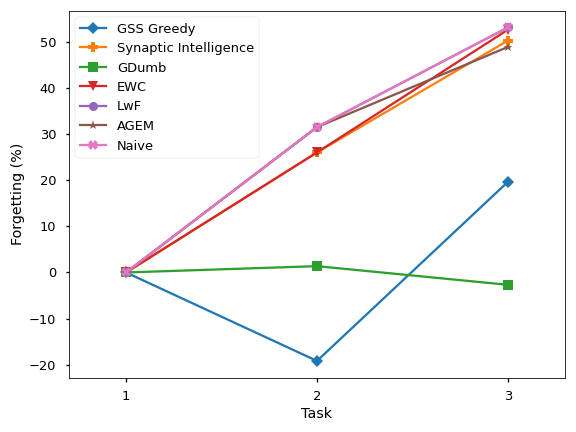}\label{fig:avg_forgetting}}
        
        \subfloat[\centering Average Accuracy]{\includegraphics[width=0.51\linewidth]{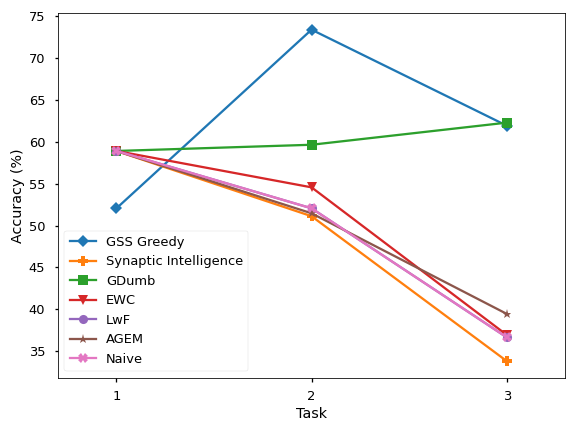}\label{fig:avg_acc}}%
   }%
   \label{fig:metric}%
    \caption{Comparison of the accuracy and forgetting metrics for all the approaches as a function of the task sequence. }%
\end{wrapfigure}





Among all of the methodologies tested, GDumb and GSS Greedy are able to retain prior knowledge from the previous tasks while learning our I/O profiling data. Interestingly, the results are different for the other methodologies used, such as EWC, LwF, Synaptic Intelligence, and AGEM. These methodologies still suffer catastrophic forgetting after Task 3 even though the Naive methodology was provided as the lowest baseline (\prettyref{fig:avg_forgetting}), which can be attributed to the fact that learning in the presence of \emph{real concept drift} is a much harder problem.

\emph{GSS Greedy} is the best methodology while training from Task 1 to Task 2. However, it suffers greatly from catastrophic forgetting after training Task 3. 
Furthermore, GSS Greedy also has the best accuracy when training a single task (\prettyref{table:results}). Consequently, after Task 3, GSS Greedy is not able to maintain its performance due to its memory size limitation and poor greedy selection of prior task knowledge. Despite its performance degradation after learning Task 3, GSS Greedy is still one of the best results we got in this experiment.
\emph{GDumb} performs best on our data by successfully retaining knowledge from Task 1 to Task 3. The adaptation of GDumb to new tasks without forgetting its previous tasks can be clearly seen in (\prettyref{fig:avg_forgetting}). 
GDumb gets the best result after training Task 3 and performed better than the other methodologies, albeit not getting the best accuracy for training a single task, in contrast to GSS Greedy.









\section{Conclusions and Future Works}

In this work, we focus on modeling applications performance inside HPC production systems which usually suffer from catastrophic forgetting due to data distribution drifts as a result of system changes over time. We highlight that the distribution drift for performance modeling follows \emph{real concept drift}
and present a continual learning-based approach to performance modeling that is able to retain prior knowledge while maintaining the ability to learn new data. We incorporate real concept drift  within the Avalanche framework to evaluate on various continual learning benchmarks. We propose a novel approach towards modeling continuous concept drift as a class incremental learning problem,  and observe that the best performing GDumb model provides a 2$\times$ improvement in accuracy over the naive approach. 

So far, we have not yet performed further hyperparameter (\prettyref{appendix:hyperparams}) optimization of the baseline model and the different continual learning methodologies used in this work. We believe that the performance of the model can be greatly improved by optimizing the hyperparameters. To date, the task id as described in \prettyref{sec:io-perf-modelling} was collected manually from the server operators. This can be automated by using the concept drift-aware modeling created by \citet{Madireddy2019-tv} which automatically identifies the points where concept drift occurs in near-real time. We hope to integrate this modeling along with our predictor to continually and adaptively learn the data in the production systems.

In the near future, we hope that the improved model can be used in the production systems. Our model needs more data with more concept drifts to learn the characteristics of applications inside the production system. Furthermore, our model can benefit other software, such as job schedulers and monitoring tools, by providing them with enough information to aid their decision-making. All in all, by integrating our model into the production system and its software, we hope that we can improve the workload optimization that spans across production systems.

\bibliographystyle{IEEEtranN}
\bibliography{references,ray-references}

\begin{thebibliography}{25}
\providecommand{\natexlab}[1]{#1}
\providecommand{\url}[1]{#1}
\csname url@samestyle\endcsname
\providecommand{\newblock}{\relax}
\providecommand{\bibinfo}[2]{#2}
\providecommand{\BIBentrySTDinterwordspacing}{\spaceskip=0pt\relax}
\providecommand{\BIBentryALTinterwordstretchfactor}{4}
\providecommand{\BIBentryALTinterwordspacing}{\spaceskip=\fontdimen2\font plus
\BIBentryALTinterwordstretchfactor\fontdimen3\font minus
  \fontdimen4\font\relax}
\providecommand{\BIBforeignlanguage}[2]{{%
\expandafter\ifx\csname l@#1\endcsname\relax
\typeout{** WARNING: IEEEtranN.bst: No hyphenation pattern has been}%
\typeout{** loaded for the language `#1'. Using the pattern for}%
\typeout{** the default language instead.}%
\else
\language=\csname l@#1\endcsname
\fi
#2}}
\providecommand{\BIBdecl}{\relax}
\BIBdecl

\bibitem[Mundt et~al.(2020)Mundt, Hong, Pliushch, and
  Ramesh]{mundt2020wholistic}
M.~Mundt, Y.~W. Hong, I.~Pliushch, and V.~Ramesh, ``A wholistic view of
  continual learning with deep neural networks: Forgotten lessons and the
  bridge to active and open world learning,'' \emph{arXiv preprint
  arXiv:2009.01797}, 2020.

\bibitem[Kasick et~al.(2010)Kasick, Tan, Gandhi, and Narasimhan]{Kasick2010-vv}
M.~P. Kasick, J.~Tan, R.~Gandhi, and P.~Narasimhan, ``{Black-Box} problem
  diagnosis in parallel file systems,'' in \emph{{FAST}}, 2010, pp. 43--56.

\bibitem[Gunawi et~al.(2018)Gunawi, Suminto, Sears, Golliher, Sundararaman,
  Lin, Emami, Sheng, Bidokhti, McCaffrey, Srinivasan, Panda, Baptist, Grider,
  Fields, Harms, Ross, Jacobson, Ricci, Webb, Alvaro, Runesha, Hao, and
  Li]{Gunawi2018-xc}
H.~S. Gunawi, R.~O. Suminto, R.~Sears, C.~Golliher, S.~Sundararaman, X.~Lin,
  T.~Emami, W.~Sheng, N.~Bidokhti, C.~McCaffrey, D.~Srinivasan, B.~Panda,
  A.~Baptist, G.~Grider, P.~M. Fields, K.~Harms, R.~B. Ross, A.~Jacobson,
  R.~Ricci, K.~Webb, P.~Alvaro, H.~B. Runesha, M.~Hao, and H.~Li, ``{Fail-Slow}
  at scale: Evidence of hardware performance faults in large production
  systems,'' \emph{ACM Trans. Storage}, vol.~14, no.~3, pp. 1--26, Oct. 2018.

\bibitem[Lockwood et~al.(2018{\natexlab{a}})Lockwood, Snyder, Wang, Byna,
  Carns, and Wright]{Lockwood2018-cl}
G.~K. Lockwood, S.~Snyder, T.~Wang, S.~Byna, P.~Carns, and N.~J. Wright, ``A
  year in the life of a parallel file system,'' in \emph{{SC18}: International
  Conference for High Performance Computing, Networking, Storage and Analysis},
  Nov. 2018, pp. 931--943.

\bibitem[Gama et~al.(2014)Gama, {\v Z}liobait{\.e}, Bifet, Pechenizkiy, and
  Bouchachia]{Gama2014-nd}
J.~Gama, I.~{\v Z}liobait{\.e}, A.~Bifet, M.~Pechenizkiy, and A.~Bouchachia,
  ``A survey on concept drift adaptation,'' \emph{ACM Comput. Surv.}, vol.~46,
  no.~4, pp. 1--37, Mar. 2014.

\bibitem[Madireddy et~al.(2019)Madireddy, Balaprakash, Carns, Latham, Lockwood,
  Ross, Snyder, and Wild]{Madireddy2019-tv}
S.~Madireddy, P.~Balaprakash, P.~Carns, R.~Latham, G.~K. Lockwood, R.~Ross,
  S.~Snyder, and S.~M. Wild, ``Adaptive learning for concept drift in
  application performance modeling,'' in \emph{Proceedings of the 48th
  International Conference on Parallel Processing}.\hskip 1em plus 0.5em minus
  0.4em\relax New York, NY, USA: ACM, Aug. 2019.

\bibitem[Thrun and Pratt(2012)]{Thrun2012-px}
S.~Thrun and L.~Pratt, \emph{\BIBforeignlanguage{en}{Learning to Learn}}.\hskip
  1em plus 0.5em minus 0.4em\relax Springer Science \& Business Media, Dec.
  2012.

\bibitem[Lesort et~al.(2021)Lesort, Caccia, and Rish]{Lesort2021-sc}
T.~Lesort, M.~Caccia, and I.~Rish, ``Understanding continual learning settings
  with data distribution drift analysis,'' Apr. 2021.

\bibitem[Cermelli et~al.(2020)Cermelli, Mancini, Rota~Bul{\`o}, Ricci, and
  Caputo]{Cermelli2020-ls}
F.~Cermelli, M.~Mancini, S.~Rota~Bul{\`o}, E.~Ricci, and B.~Caputo, ``Modeling
  the background for incremental learning in semantic segmentation,'' in
  \emph{2020 {IEEE/CVF} Conference on Computer Vision and Pattern Recognition
  ({CVPR})}, Jun. 2020, pp. 9230--9239.

\bibitem[Abdelsalam et~al.(2021)Abdelsalam, Faramarzi, Sodhani, and
  Chandar]{Abdelsalam2021-ne}
M.~Abdelsalam, M.~Faramarzi, S.~Sodhani, and S.~Chandar, ``{IIRC}: Incremental
  {Implicitly-Refined} classification,'' in \emph{2021 {IEEE/CVF} Conference on
  Computer Vision and Pattern Recognition ({CVPR})}, Jun. 2021, pp.
  11\,033--11\,042.

\bibitem[Lockwood et~al.(2018{\natexlab{b}})Lockwood, Wright, Snyder, Carns,
  Brown, and Harms]{Lockwood2018-aw}
G.~K. Lockwood, N.~J. Wright, S.~Snyder, P.~Carns, G.~Brown, and K.~Harms,
  ``\BIBforeignlanguage{en}{{TOKIO} on {ClusterStor}: Connecting standard tools
  to enable holistic {I/O} performance analysis},'' Lawrence Berkeley National
  Lab. (LBNL), Berkeley, CA (United States), Tech. Rep., Jan. 2018.

\bibitem[Carns et~al.(2011)Carns, Harms, Allcock, Bacon, Lang, Latham, and
  Ross]{Carns2011-vc}
P.~Carns, K.~Harms, W.~Allcock, C.~Bacon, S.~Lang, R.~Latham, and R.~Ross,
  ``Understanding and improving computational science storage access through
  continuous characterization,'' \emph{ACM Trans. Storage}, vol.~7, no.~3, pp.
  1--26, Oct. 2011.

\bibitem[{Garlick} and {Morrone}()]{Garlick_undated-yd}
{Garlick} and {Morrone}, ``Lustre monitoring tools,''
  \url{https://github.com/LLNL/lmt}.

\bibitem[Yoo et~al.(2003)Yoo, Jette, and Grondona]{Yoo2003-dx}
A.~B. Yoo, M.~A. Jette, and M.~Grondona, ``{SLURM}: Simple linux utility for
  resource management,'' in \emph{Job Scheduling Strategies for Parallel
  Processing}.\hskip 1em plus 0.5em minus 0.4em\relax Springer Berlin
  Heidelberg, 2003, pp. 44--60.

\bibitem[French(1999)]{French1999-sv}
R.~M. French, ``\BIBforeignlanguage{en}{Catastrophic forgetting in
  connectionist networks},'' \emph{\BIBforeignlanguage{en}{Trends Cogn. Sci.}},
  vol.~3, no.~4, pp. 128--135, Apr. 1999.

\bibitem[Lomonaco et~al.(2021)Lomonaco, Pellegrini, Cossu, Carta, Graffieti,
  Hayes, Lange, Masana, Pomponi, van~de Ven, Mundt, She, Cooper, Forest,
  Belouadah, Calderara, Parisi, Cuzzolin, Tolias, Scardapane, Antiga, Amhad,
  Popescu, Kanan, van~de Weijer, Tuytelaars, Bacciu, and Maltoni]{avalanche}
V.~Lomonaco, L.~Pellegrini, A.~Cossu, A.~Carta, G.~Graffieti, T.~L. Hayes,
  M.~D. Lange, M.~Masana, J.~Pomponi, G.~van~de Ven, M.~Mundt, Q.~She,
  K.~Cooper, J.~Forest, E.~Belouadah, S.~Calderara, G.~I. Parisi, F.~Cuzzolin,
  A.~Tolias, S.~Scardapane, L.~Antiga, S.~Amhad, A.~Popescu, C.~Kanan,
  J.~van~de Weijer, T.~Tuytelaars, D.~Bacciu, and D.~Maltoni, ``Avalanche: an
  end-to-end library for continual learning,'' in \emph{Proceedings of IEEE
  Conference on Computer Vision and Pattern Recognition}, ser. 2nd Continual
  Learning in Computer Vision Workshop, 2021.

\bibitem[Paszke et~al.(2019)Paszke, Gross, Massa, Lerer, Bradbury, Chanan,
  Killeen, Lin, Gimelshein, Antiga, and {Others}]{Paszke2019-cr}
A.~Paszke, S.~Gross, F.~Massa, A.~Lerer, J.~Bradbury, G.~Chanan, T.~Killeen,
  Z.~Lin, N.~Gimelshein, L.~Antiga, and {Others}, ``Pytorch: An imperative
  style, high-performance deep learning library,'' \emph{Adv. Neural Inf.
  Process. Syst.}, vol.~32, 2019.

\bibitem[Masana et~al.(2020)Masana, Liu, Twardowski, Menta, Bagdanov, and
  van~de Weijer]{Masana2020-ln}
M.~Masana, X.~Liu, B.~Twardowski, M.~Menta, A.~D. Bagdanov, and J.~van~de
  Weijer, ``Class-incremental learning: survey and performance evaluation on
  image classification,'' Oct. 2020.

\bibitem[Kirkpatrick et~al.(2017)Kirkpatrick, Pascanu, Rabinowitz, Veness,
  Desjardins, Rusu, Milan, Quan, Ramalho, Grabska-Barwinska, Hassabis, Clopath,
  Kumaran, and Hadsell]{kirkpatrick_2017_Overcoming}
J.~Kirkpatrick, R.~Pascanu, N.~Rabinowitz, J.~Veness, G.~Desjardins, A.~A.
  Rusu, K.~Milan, J.~Quan, T.~Ramalho, A.~Grabska-Barwinska, D.~Hassabis,
  C.~Clopath, D.~Kumaran, and R.~Hadsell, ``Overcoming catastrophic forgetting
  in neural networks,'' \emph{Proceedings of the National Academy of Sciences
  of the United States of America}, vol. 114, no.~13, pp. 3521--3526, 2017.

\bibitem[Zenke et~al.(2017)Zenke, Poole, and Ganguli]{zenke2017continual}
F.~Zenke, B.~Poole, and S.~Ganguli, ``Continual learning through synaptic
  intelligence,'' in \emph{Proceedings of the 34th International Conference on
  Machine Learning}, vol.~70.\hskip 1em plus 0.5em minus 0.4em\relax JMLR. org,
  2017, pp. 3987--3995.

\bibitem[Li and Hoiem(2017)]{li2017learning}
Z.~Li and D.~Hoiem, ``Learning without forgetting,'' \emph{IEEE transactions on
  pattern analysis and machine intelligence}, vol.~40, no.~12, pp. 2935--2947,
  2017.

\bibitem[Chaudhry et~al.(2018)Chaudhry, Ranzato, Rohrbach, and
  Elhoseiny]{chaudhry2018efficient}
A.~Chaudhry, M.~Ranzato, M.~Rohrbach, and M.~Elhoseiny, ``Efficient lifelong
  learning with a-gem,'' \emph{arXiv preprint arXiv:1812.00420}, 2018.

\bibitem[Aljundi et~al.(2019)Aljundi, Lin, Goujaud, and
  Bengio]{aljundi2019gradient}
R.~Aljundi, M.~Lin, B.~Goujaud, and Y.~Bengio, ``Gradient based sample
  selection for online continual learning,'' \emph{Advances in neural
  information processing systems}, vol.~32, 2019.

\bibitem[Prabhu et~al.(2020)Prabhu, Torr, and Dokania]{prabhu2020gdumb}
A.~Prabhu, P.~H. Torr, and P.~K. Dokania, ``Gdumb: A simple approach that
  questions our progress in continual learning,'' in \emph{European conference
  on computer vision}.\hskip 1em plus 0.5em minus 0.4em\relax Springer, 2020,
  pp. 524--540.

\bibitem[Mai et~al.(2021)Mai, Li, Jeong, Quispe, Kim, and Sanner]{Mai2021-er}
Z.~Mai, R.~Li, J.~Jeong, D.~Quispe, H.~Kim, and S.~Sanner, ``Online continual
  learning in image classification: An empirical survey,'' Jan. 2021.

\end{thebibliography}

\newpage

\appendix

\section{Appendix}



\subsection{Hyperparameters}

In Table \ref{tab:hyperparams}, we defined the hyperparameters of neural network models, training and evaluation configurations, and continual learning approaches parameter that we used through the experiments.

\label{appendix:hyperparams}

\newlength{\oldintextsep}
\setlength{\oldintextsep}{\intextsep}

\setlength\intextsep{0pt}
\begin{table}[ht!]
  \caption{Hyperparameters}
  \centering
  \vspace{0.1in}
  \label{tab:hyperparams}
  \begin{threeparttable}[t!]
    \resizebox{0.45\textwidth}{!}
     {
      \begin{tabular}{llc}
        \toprule
        & Parameters & Value \\
        \midrule
        \midrule
        General & Epochs
            & 60 \\
        & Training Batch Size       
            & 4 \\
        & Test Batch Size       
            & 4 \\
        & Learning Rate
            & 0.001 \\
        & Optimizer 
            & Adam \\
        & Hidden Layers
            & 3 \\
        & Hidden Size
            & 400 \\
        \midrule
        Synaptic Intelligence & Lambda & 1.0 \\
        & Eps & \( \num{1e-7} \) \\
        \midrule
        EWC & Mode & Separate \\
        & Lambda & \( 0.5 \) \\
        \midrule
        LwF & Alpha &  \( 1.0 \) \\
        & Temperature & \( 2.0 \) \\
        \midrule
        AGEM & Patterns Per Exp & 2 \\
        \midrule
        GSS Greedy & Mem Size & 5000 \\
        \midrule
        GDumb & Mem Size & 5000 \\
        \midrule
        \bottomrule
      \end{tabular}
    }
    \end{threeparttable}
\end{table}

\end{document}